\documentclass[aps,prx,preprint,showpacs,superscriptaddress]{revtex4-1}
\usepackage{latexsym}
\usepackage{amssymb}
\usepackage{graphicx}
\usepackage{amsmath}
\usepackage{bm}
\usepackage[colorlinks,
          linkcolor=black,
            citecolor=black,
            urlcolor=blue
           ]{hyperref}
\usepackage{verbatim}
\usepackage{mathrsfs}
\usepackage{extarrows}
\usepackage{comment}
\usepackage{mathtools,slashed}

\begin{document}

\title{An exact solution in Markov decision process with multiplicative rewards as a general framework}

\author{Yuan Yao}
\thanks{Both authors contributed equally to the work.}
\email{smartyao@issp.u-tokyo.ac.jp}
\affiliation{Institute for Solid State Physics, University of Tokyo, Kashiwa, Chiba 277-8581, Japan}
\affiliation{Condensed Matter Theory Laboratory, RIKEN CPR, Wako, Saitama 351-0198, Japan}
\author{Xiaolin Sun}
\thanks{Both authors contributed equally to the work.}
\email{1155022058@edu.k.u-tokyo.ac.jp}
\affiliation{Graduate School of Frontier Sciences, The University of Tokyo, Kashiwa, Chiba 277-8581, Japan}

\begin{abstract}
We develop an exactly solvable framework of Markov decision process with a finite horizon, and continuous state and action spaces. 
We first review the exact solution of conventional linear quadratic regulation with a linear transition and a Gaussian noise, whose optimal policy does not depend on the Gaussian noise, which is an undesired feature in the presence of significant noises. 
It motivates us to investigate exact solutions which depend on noise. 
To do so, we generalize the reward accumulation to be a general binary commutative and associative operation. 
By a new multiplicative accumulation,
we obtain an exact solution of optimization assuming linear transitions with a Gaussian noise and the optimal policy is noise dependent in contrast to the additive accumulation.
Furthermore, 
we also show that the multiplicative scheme is a general framework that covers the additive one with an arbitrary precision, which is a model-independent principle.
\end{abstract}


\maketitle
\section{Introduction}
Markov decision process (MDP) is stochastic control process in a discrete time~\cite{Puterman:2014aa} and it has played an essential role in studying optimizations investigated by dynamic programming~\cite{Bellman:2015aa} and reinforcement learning~\cite{Sutton:1998aa}. 
They are extensively applied in many areas, such as automatic control and robotics. 
A MDP with a finite horizon $T$, 
which is of interest of this paper, 
is a tuple $(S,A,\{P_{a}(s',s)\},R)$, where $S$ is a state space and $A$ is the set of action, both of which are continuous. Here $\{P_{a}(s',s)\}$ is the state transition probability density to $s'\in S$ from $s\in S$ exerted by the action $a\in A$ and it satisfies the Markov property. 
Moreover, $R^t: S\times A\rightarrow\mathbb{R}$ is called the reward function defined at each time step $t$, and if the process is in some state $s\in S$ and the action $a$ is chosen, a corresponding $R^t(s,a)$ is immediately accumulated. 
The task of the optimization control is to select an optimal policy $\pi_t^*$ at the time $t\in[0,T]$ from the set of policies \{$\pi:S\rightarrow A$\} to maximize the expectation of total accumulated rewards $\mathcal{R}^t$ in the unknown future $[t,T]$. 
Conventionally, $\mathcal{R}^t$ is simply the summation, denoted by $\mathcal{R}_+^t$, of the rewards obtained: 
\begin{eqnarray}
\label{Areward}
\mathcal{R}_+^t(s_t,a_t;\cdots;s_T,a_T)= R^t(s_t,a_t)+R^{t+1}(s_{t+1},a_{t+1})+\cdots+R^{T}(s_T,a_T).
\end{eqnarray}
By the definition of the optimality, the optimal policy, denoted by $\pi^*_{t,+}$ to emphasize the accumulation way as the addition, takes the form as
\begin{eqnarray}
\label{optimization}
\pi^*_{t,+}(s_t)=\text{argmax}_{a_t}\text{max}_{a_{t+1},\cdots,a_{T}}\mathbb{E}^{(t+1)}\left[\mathcal{R}_+^t(s_t,a_t;s_{t+1},a_{t+1};\cdots;s_T,a_T)\right], 
\end{eqnarray}
where the expectation $\mathbb{E}^{(t+1)}$ is taken on the distributions $\left\{s_{k}\sim P_{a_{k-1}}(s_k,s_{k-1})|t<k< T\right\}$. 

If a linear transition with Gaussian noise is assumed and a linear quadratic reward is chosen, 
such an optimization can be exactly solved. 
As we will briefly review later, 
the solution of the optimal policy $\pi^*_{t,+}$ does not depend on the noise of the linear transition~\cite{Kwakernaak:1972aa} (or c.f. Eq.~(\ref{aLQR})), which implies that the noise plays a completely trivial role there.
However, the noise is important in real systems and noise dependences in the solution of optimal policy can probably indicate whether the model in the consideration is reasonable or not.
Thus, it is interesting to extend this solution or discover other exactly solvable cases where optimal policies are noise dependent and such a control is expected to be advantageous in the presence of significant noise. 
Furthermore,
the generality of the additive way to accumulate rewards still remains open and we are interested in a more general framework than the additive accumulation, i.e., multiplicative rewards, which is the other main goal of the current paper with a positive statement.
Our proposal of such a multiplicative scheme, 
by the definition given later,
should be distinguished from the similar terminology of multiplicative MDP~\cite{Howard:1972aa,Sladky:1976aa,Rothblum:1984aa,Borkar:2002aa,Kallenberg:2011aa,Osogami:2012aa,White:2018aa,Freitas:2018aa,Bertsekas:2019aa} where the product means the additive rewards being multiplied by a one-period transition matrices.

This paper is organized as follows.
In Sec.~\ref{sec_axiom}, we propose a general paradigm of reward accumulations to generalize the concept of the additive accumulation.
We review the exact solution under the additive rewards and derive one of our main results as an exactly solvable noise dependent optimization under a multiplicative reward in Sec.~\ref{sec_add}.
Shed light by this result,
we show that our proposal of the multiplicative scheme of reward accumulation is actually a general framework, even model-independently, in Sec.~\ref{sec_mul}, followed by the conclusion in Sec.~\ref{sec_con}.

\section{An axiomatic approach to reward accumulations}
\label{sec_axiom}
In this work to construct other exact solutions whose optimal policy is noise dependent, we will investigate distinct ways that the rewards $\{R^t\}$ are accumulated 
and first consider a generalization in the accumulation and optimization as
\begin{eqnarray}
\label{optimization_1}
\pi^*_t(s_t)=\text{argmax}_{a_t}\text{max}_{a_{t+1},\cdots,a_{T}}\mathbb{E}^{(t+1)}\left[\mathcal{R}^t(s_t,a_t;s_{t+1},a_{t+1};\cdots;s_T,a_T)\right], \\
\label{Areward_1}
\mathcal{R}^t(s_t,a_t;\cdots;s_T,a_T)= R^t(s_t,a_t)\oplus R^{t+1}(s_{t+1},a_{t+1})\oplus\cdots\oplus R^{T}(s_T,a_T),
\end{eqnarray}
where the binary operation $\oplus:\mathbb{R}\times\mathbb{R}\rightarrow\mathbb{R}$ denotes a general accumulating way. 
We require that such a general accumulation satisfies the following two conditions: 
\begin{eqnarray}
r_1\oplus r_2=r_2\oplus r_1;\,\,r_1\oplus (r_2\oplus r_3)=(r_1\oplus r_2)\oplus r_3. 
\end{eqnarray}
They implies that, respectively, the order of the rewards and the order of accumulating are irrelevant~\footnote{However, for the infinite horizon, the existence of an analogous discount factor will invalidate the commutativity}.
Clearly the conventional accumulation is $\oplus=+:\mathbb{R}\times\mathbb{R}\rightarrow\mathbb{R}$, i.e. the traditional summation of two real numbers, which trivially satisfies the conditions above. 
One another simple and natural choice is the multiplication $\oplus=\cdot$, which will be our focus: 
\begin{eqnarray}
\mathcal{R}_\cdot^t(s_t,a_t;s_{t+1},a_{t+1};\cdots)=R^t(s_t,a_t)\cdot R^{t+1}(s_{t+1},a_{t+1})\cdots, 
\end{eqnarray}
where the subscript ``${}_\cdot$'' of $\mathcal{R}_\cdot^t$ denotes the multiplication.

At first glance, the multiplicative accumulation is similar to the conventional additive one since it can be transformed to be a summation form as
\begin{eqnarray}
\label{log}
\ln\mathcal{R}_\cdot^t=\ln R^t(s_t,a_t)+\ln R^{t+1}(s_{t+1},a_{t+1})+\cdots. 
\end{eqnarray}
Nevertheless, the essential distinction is in the optimal policy (\ref{optimization_1}) because for general distribution $\mathbb{E}[\ln(\cdot)]\neq \ln\mathbb{E}[\cdot]$ unless the process is completely deterministic due to Jensen's inequality~\cite{Rudin:2006aa}.
It makes the multiplicative way and the additive way quantitatively different in nature. 
As just mentioned, the optimization of multiplicatively accumulated rewards is equivalent to that of additive ones when the uncertainty in $P_{a}(s',s)$ is absent. 
This observation will be a useful checker later. 

\section{Additive and multiplicative rewards}
\label{sec_add}
In this section, we will first review an exact solution to MDP for the additive rewards $\oplus=+$~\cite{Kwakernaak:1972aa}. 
Then we will derive an analogous exactly solvable MDP with a multiplicative reward $\oplus=\cdot$ as our main result. 
\subsection{Linear transition with a Gaussian noise}
In the following discussion, 
we assume the following linear transition for $t=1,\cdots,T$
\begin{eqnarray}
\label{LT}
\left\{\begin{array}{l}s_{t}=A_{t-1}s_{t-1}+B_{t-1}a_{t-1}+w_{t-1};\\
w_{t-1}\sim\mathcal{N}(0,\Sigma_{t-1}),\end{array}\right.
\end{eqnarray}
where the Gaussian distribution $\mathcal{N}(\mu,\Sigma)$ takes the form of
\begin{eqnarray}
\label{white_noise}
p(x;\mu,\Sigma)=\frac{1}{\sqrt{|2\pi\Sigma|}}\exp\left[-\frac{1}{2}(x-\mu)^\intercal\Sigma^{-1}(x-\mu)\right]. 
\end{eqnarray}

\subsection{Additive linear quadratic rewards}
Let us first review the result of $\oplus=+$ with the following quadratic rewards together with the linear transition (\ref{LT}) above called linear quadratic regulator (LQR)~\cite{Kwakernaak:1972aa}:
\begin{eqnarray}
\label{LQR}
R_\text{LQR}^{t}(s_t,a_t)=-s_t^\intercal U_ts_t-a_t^\intercal W_ta_t, 
\end{eqnarray}
where $U_t$ and $W_t$ are positive definite matrices. 
The choice $\oplus=+$ means the cumulative rewards is the following summation:
\begin{eqnarray}
\mathcal{R}_{+,\text{LQR}}^t(s_t,a_t;\cdots;s_T,a_T)= \sum_{k=t}^TR_\text{LQR}^k(s_k,a_k)=\sum_{k=t}^T-s_k^\intercal U_ks_k-a_k^\intercal W_ka_k.
\end{eqnarray}
Then the optimal policy can be obtained by Eq.~(\ref{optimization}) as
\begin{eqnarray}
\label{aLQR}
\pi_{t,+\text{ LQR}}^*(s_t)=\left[(W_t-B^\intercal_t\Phi_{t+1\text{ LQR}}B_t)^{-1}B_t\Phi_{t+1\text{ LQR}}A_t\right]s_t, 
\end{eqnarray}
where $\Phi_{t\text{ LQR}}$ is \emph{backward} updated by $\Phi_{t+1\text{ LQR}}$ through the following Riccati equation: 
\begin{eqnarray}
\label{riccati}
\Phi_{t\text{ LQR}}=A^\intercal_t\left[\Phi_{t+1\text{ LQR}}+\Phi_{t+1\text{ LQR}}B_t(W_t-B_t^\intercal\Phi_{t+1\text{ LQR}}B_t)^{-1}B_t\Phi_{t+1\text{ LQR}}\right]A_t-U_t, 
\end{eqnarray}
with the initialization as
\begin{eqnarray}
\Phi_{T+1\text{ LQR}}=0. 
\end{eqnarray}
Although we will not re-derive this exact solution, 
it should be noted that the optimal policy (\ref{aLQR}) is completely independent of the noise (\ref{aLQR}) in linear transitions.
This property enables us to apply the LQR even without measuring the covariance matrices $\Sigma$.
It is also reflected in the fact that the updating rule (\ref{riccati}) is the Riccati equation for the deterministic optimal control with LQR.

\subsection{Multiplicative linear exponentiated quadratic rewards}
In this part, we will investigate the case of $\oplus=\cdot$ in Eq.~(\ref{Areward_1}), i.e. the multiplicative rewards: 
\begin{eqnarray}
\mathcal{R}_\cdot^t(s_t,a_t;\cdots;s_T,a_T)=\prod_{k=t}^TR^k(s_k,a_k),
\end{eqnarray}
with the following exponentiated quadratic reward
\begin{eqnarray}
R^k(s_ka_k)=\exp\left(-s_k^\intercal U_ks_k-a_k^\intercal W_ka_k\right)
\end{eqnarray}
which implies
\begin{eqnarray}
\label{reward_m}
\mathcal{R}_\cdot^t(s_t,a_t;\cdots;s_T,a_T)=\exp\left(\sum_{k=t}^T-s_k^\intercal U_ks_k-a_k^\intercal W_ka_k\right).
\end{eqnarray}
We recall from Eq.~(\ref{optimization_1}) that 
\begin{eqnarray}
\label{a*}
\pi^*_{t,\cdot}(s_t)=\text{argmax}_{a_t}\text{max}_{a_{t+1},\cdots,a_{T}}\mathbb{E}^{(t+1)}\left[\exp\left(\sum_{k=t}^T-s_k^\intercal U_ks_k-a_k^\intercal W_ka_k\right)\right].
\end{eqnarray}
It is natural to define the value function as the maximization in Eq.~(\ref{a*}):
\begin{eqnarray}
\label{value}
V^*_t(s_t)=\text{max}_{a_t}\text{max}_{a_{t+1},\cdots,a_{T}}\mathbb{E}^{(t+1)}\left[\mathcal{R}_\cdot^t(s_t,a_t;s_{t+1},a_{t+1};\cdots;s_T,a_T)\right],
\end{eqnarray}
Let us first consider $\pi_{T,\cdot}^*(s_T)$ from $V_T^*(s_T)$ in Eq.~(\ref{value}) since the world ends at $T$: 
\begin{eqnarray}
\label{ind_0}
V_T^*(s_T)&=&\text{max}_{a_T}[\mathcal{R}_\cdot^T(s_T,a_T)]\nonumber\\
&=&\text{max}_{a_T}[\exp\left(-s_T^\intercal U_Ts_T-a_T^\intercal W_Ta_T\right)]\nonumber\\
&=&\exp\left(-s_T^\intercal U_Ts_T\right), 
\end{eqnarray}
with $\pi_T^*=0$ due to the positive-definiteness of the matrix $W_T$. 
Therefore, observing Eq.~(\ref{ind_0}), we would set the induction assumption as
\begin{eqnarray}
\label{ind_ass}
V_{t+1}^*(s_{t+1})\stackrel{?}{=}\frac{1}{D_{t+1}}\exp(s^\intercal_{t+1}\Phi_{t+1}s_{t+1}), 
\end{eqnarray}
for some to-be-determined matrices $\Phi_{t+1}$ and number $D_{t+1}$ independent on $s_{t+1}$. 
Our main task is to prove that $V_t^*(s_t)$ is of the exactly the same form with some $\Phi_t$ and $D_t$ derived from $\Phi_{t+1}$ and $D_{t+1}$. 
By definitions in Eqs.~(\ref{reward_m},\ref{value}) and the induction assumption (\ref{ind_ass}), 
\begin{eqnarray}
\label{value_1}
V^*_t(s_t)&=&\text{max}_{a_t}\left\{\exp(-s_t^\intercal U_ts_t-a_t^\intercal W_ta_t)\cdot\text{max}_{a_{t+1},\cdots,a_{T}}\mathbb{E}^{(t+1)}\left[\mathcal{R}_\cdot^{t+1}(s_{t+1},a_{t+1};\cdots;s_T,a_T)\right]\right\}\nonumber\\
&=&\exp(-s_t^\intercal U_ts_t)\text{max}_{a_t}\left[\exp(-a_t^\intercal W_ta_t)\mathbb{E}_{s_{t+1}\sim\mathcal{N}(A_ts_t+B_ta_t,\Sigma_t)}V_{t+1}^*(s_{t+1})\right]\nonumber\\
&=&\frac{\exp(-s_t^\intercal U_ts_t)}{D_{t+1}}\text{max}_{a_t}\left\{\exp(-a_t^\intercal W_ta_t)\mathbb{E}_{s_{t+1}\sim\mathcal{N}(A_ts_t+B_ta_t,\Sigma_t)}\left[\exp(s^\intercal_{t+1}\Phi_{t+1}s_{t+1})\right]\right\}, 
\end{eqnarray}
where $\mathbb{E}_{s_{t+1}\sim\mathcal{N}(A_ts_t+B_ta_t,\Sigma_t)}$ precisely means the sampling of $s_{t+1}$ by the linear transition with a Gaussian white distribution as in Eq.~(\ref{LT}). 
Then we extend out such a Gaussian integration in Eq.~(\ref{value_1}):
\begin{eqnarray}
\label{value_2}
V^*_t(s_t)&=&\exp(-s_t^\intercal U_ts_t)\frac{1}{D_{t+1}}\text{max}_{a_t}\left\{\exp(-a_t^\intercal W_ta_t)\int d\vec{s}_{t+1}\sqrt{|\Sigma^{-1}_t/2\pi|}\right.\nonumber\\
&&\left.\exp\left[-\frac{1}{2}(s_{t+1}-A_ts_t-B_ta_t)^\intercal\Sigma_t^{-1}(s_{t+1}-A_ts_t-B_ta_t)\right]\exp(s^\intercal_{t+1}\Phi_{t+1}s_{t+1})\right\}\nonumber\\
&=&\frac{1}{D_{t+1}}\sqrt{\frac{|\Sigma_t^{-1}|}{|\Sigma_t^{-1}-2\Phi_{t+1}|}}\exp(-s_t^\intercal U_ts_t)\nonumber\\
&&\text{max}_{a_t}\left\{\exp\left[-a^\intercal_tW_ta_t+(A_ts_t+B_ta_t)^\intercal\Omega_{t+1}(A_ts_t+B_ta_t)\right]\right\}, 
\end{eqnarray}
where
\begin{eqnarray}
\Omega_{t+1}&\equiv&\Sigma_t^{-1}\left(\Sigma_t^{-1}-2\Phi_{t+1}\right)^{-1}\Phi_{t+1}. 
\end{eqnarray}
Therefore, we obtain one of the main results as: 
\begin{eqnarray}
\pi^*_{t,\cdot}(s_t)&=&\text{argmax}_{a_t}\left\{\exp\left[-a^\intercal_tW_ta_t+(A_ts_t+B_ta_t)^\intercal\Omega_{t+1}(A_ts_t+B_ta_t)\right]\right\}\nonumber\\
&=&\left[(W_t-B^\intercal_t\Omega_{t+1}B_t)^{-1}B_t\Omega_{t+1}A_t\right]s_t, 
\end{eqnarray}
which is put into Eq.~(\ref{value_2}) to derive that
\begin{eqnarray}
\label{value_3}
V^*_t(s_t)&=&\frac{1}{D_{t+1}}\sqrt{\frac{|\Sigma_t^{-1}|}{|\Sigma_t^{-1}-2\Phi_{t+1}|}}\exp\left[-s_t^\intercal (U_t-A^\intercal_t\Omega_{t+1}A_t)s_t\right]\nonumber\\
&&\exp\left[s^\intercal_tA_t^\intercal\Omega_{t+1}B_t(W_t-B_t^\intercal\Omega_{t+1}B_t)^{-1}B_t^\intercal\Omega_{t+1}A_ts_t\right]\nonumber\\
&\equiv&\frac{1}{D_{t}}\exp\left(s^\intercal_t\Phi_ts_t\right), 
\end{eqnarray}
with
\begin{eqnarray}
\frac{1}{D_t}\equiv \frac{1}{D_{t+1}}\sqrt{\frac{|\Sigma_t^{-1}|}{|\Sigma_t^{-1}-2\Phi_{t+1}|}}
\end{eqnarray}
and the following updating rule:
\begin{eqnarray}
\Phi_t\equiv A^\intercal_t\left[\Omega_{t+1}+\Omega_{t+1}B_t(W_t-B_t^\intercal\Omega_{t+1}B_t)^{-1}B_t\Omega_{t+1}\right]A_t-U_t. 
\end{eqnarray}
Indeed, we have proven the induction step that $V_t^*(s_t)$ also precisely takes the exponentiated quadratic form with $\Phi_t$ and $D_t$ independent on $s_t$. 
From Eq.~(\ref{ind_0}), 
we obtain the backward initialization as
\begin{eqnarray}
\left\{\begin{array}{l}\Phi_{T}=-U_T;\\
\frac{1}{D_T}=1,\end{array}\right.
\end{eqnarray}
in addition to $\pi^*_T=0$ from the last line of Eq.~(\ref{ind_0}). 
Of course, we can also embed $\pi^*_T=0$ into the updating rules by artificially extending horizon to $(T+1)$ by
\begin{eqnarray}
\label{compact_ini}
\left\{\begin{array}{l}\Phi_{T+1}=0;\\
\frac{1}{D_{T+1}}=1.\end{array}\right.
\end{eqnarray}
In a short summary, with the initialization (\ref{compact_ini}), 
\begin{eqnarray}
\label{policy_m}
\pi^*_{t,\cdot}(s_t)&=&\left[(W_t-B^\intercal_t\Omega_{t+1}B_t)^{-1}B_t\Omega_{t+1}A_t\right]s_t, \\
\label{omega_m}
\Omega_{t+1}&\equiv&\Sigma_t^{-1}\left(\Sigma_t^{-1}-2\Phi_{t+1}\right)^{-1}\Phi_{t+1};
\end{eqnarray}
with the updating rule for $\Omega_{t+1}$ which is determined by the updating of $\Phi_{t+1}$: 
\begin{eqnarray}
\label{update_m}
\Phi_t= A^\intercal_t\left[\Omega_{t+1}+\Omega_{t+1}B_t(W_t-B_t^\intercal\Omega_{t+1}B_t)^{-1}B_t\Omega_{t+1}\right]A_t-U_t.
\end{eqnarray}

\subsection{The deterministic case: $\Sigma_k\rightarrow0$}
\label{noise_free}
It is noted that our exponentiated reward (\ref{reward_m}) is related to the additive one in Eq.~(\ref{LQR}) by the logarithm (\ref{log}). 
Thus, by Jensen's inequality, it is a consistency check that the optimal policy (\ref{policy_m}) should be reduced to the policy (\ref{aLQR}) (that is noise-independent) in the noise-free limit $\{\Sigma_k\rightarrow 0\}$, 
\begin{eqnarray}
\lim_{\{\Sigma_k\rightarrow0\}}\pi_{t,\cdot}^*=\lim_{\{\Sigma_k\rightarrow0\}}\pi_{t,+\text{LQR}}^*=\pi_{t,+\text{LQR}}^*. 
\end{eqnarray}
which is indeed the case since $\Omega_{t+1}\rightarrow\Phi_{t+1}$ in Eq.~(\ref{omega_m}) in such a limit and the updating rule~(\ref{update_m}) becomes the standard Riccati equation~(\ref{riccati}). 
We will see that this reduction under a special noise limit to the policy under additive reward reflects a general principle that the scheme of multiplicative reward is more general than the additive scheme independently of the model.

\section{Multiplicative scheme as a general framework}
\label{sec_mul}

In the discussions above, 
we propose a new multiplicative reward accumulating way other than the additive one.
However, 
it appears that whether the multiplicative or the additive one works better for the practical sake strongly depends on the system in the real world.
Actually,
we will prove that the multiplicative approach is a general framework, i.e., any optimal policy obtained by a certain additive reward function can be approximated by the policy obtained by a multiplicative reward with an arbitrary precision.
\subsection{Scaling invariance}
To address the issue above,
let us observe the optimal policy in Eqs.~(\ref{policy_m},\ref{omega_m},\ref{update_m}) rewritten below:
\begin{eqnarray}
\label{policy_m_1}
\pi^*_{t,\cdot}(s_t)[\{W_t,U_t,\Sigma_t\}]&=&\left[(W_t-B^\intercal_t\Omega_{t+1}B_t)^{-1}B_t\Omega_{t+1}A_t\right]s_t, \\
\Omega_{t+1}&\equiv&\Sigma_t^{-1}\left(\Sigma_t^{-1}-2\Phi_{t+1}\right)^{-1}\Phi_{t+1};
\end{eqnarray}
with the updating rule for $\Omega_{t+1}$ which is determined by the updating of $\Phi_{t+1}$: 
\begin{eqnarray}
\Phi_t= A^\intercal_t\left[\Omega_{t+1}+\Omega_{t+1}B_t(W_t-B_t^\intercal\Omega_{t+1}B_t)^{-1}B_t\Omega_{t+1}\right]A_t-U_t.
\end{eqnarray}
We have made explicit the parameter dependence on $\{W_t\}$,$\{U_t\}$ and $\Sigma_t$ in Eq.~(\ref{policy_m_1}).
It is straightforward to prove the following scaling invariance:
\begin{eqnarray}
\pi^*_{t,\cdot}(s_t)[\{W_t,U_t,\Sigma_t\}]&=&\pi^*_{t,\cdot}(s_t)[\{\kappa W_t,\kappa U_t,\kappa^{-1}\Sigma_t\}],
\end{eqnarray}
which can be also rearranged into the following form
\begin{eqnarray}
\label{scaling}
\pi^*_{t,\cdot}(s_t)[\{W_t,U_t,\kappa\Sigma_t\}]&=&\pi^*_{t,\cdot}(s_t)[\{\kappa W_t,\kappa U_t,\Sigma_t\}].
\end{eqnarray}
On the other hand,
the noise-free limit in Sec.~\ref{noise_free} implies that
\begin{eqnarray}
\lim_{\kappa\rightarrow0^+}\pi^*_{t,\cdot}(s_t)[\{W_t,U_t,\kappa\Sigma_t\}]&=&\pi^*_{t,+}(s_t)[\{W_t,U_t,\Sigma_t\}],
\end{eqnarray}
which means, by the scaling invariance~(\ref{scaling}), that
\begin{eqnarray}
\lim_{\kappa\rightarrow0^+}\pi^*_{t,\cdot}(s_t)[\{\kappa W_t,\kappa U_t\}]&=&\pi^*_{t,+}(s_t)[\{W_t,U_t\}],
\end{eqnarray}
where we have removed the redundant (the same) noise dependence.

Therefore, 
the scaling invariance~(\ref{scaling}) ensures that the optimal policy under the additive reward~(\ref{LQR}) can be approached by including a sufficiently small scaling coefficient $\kappa$ in the multiplicative reward~(\ref{reward_m}).

Furthermore,
this phenomenon is not reward-function dependent or even model dependent.
For any additive upper-bouded reward function $R^t_+$ at the time slice $t$ with its accumulation $\mathcal{R}^t_+\equiv\sum_{k\geq t}R^k_+$,
we can define the following multiplicative reward with its accumulation:
\begin{eqnarray}
R^t_\cdot\equiv\exp\left(\kappa R^t_+\right)\text{ with }\mathcal{R}^t_\cdot\equiv\prod_{k\geq t}R^t_\cdot,
\end{eqnarray}
whose optimal policy denoted by $\pi^*_{t,\cdot}$ can be reduced to the optimal policy derived by $\{R^t_+\}$ in the limit:
\begin{eqnarray}
\lim_{\kappa\rightarrow0^+}\pi^*_{t,\cdot}[\{R^t_\cdot\}]=\pi^*_{t,+}[\{R^t_+\}].
\end{eqnarray}
It is because of the Taylor expansion that
\begin{eqnarray}
R^t_\cdot=1+\kappa R^t_++O(\kappa^2),
\end{eqnarray}
where the constant term does not contribute to the optimal policy and the higher order term $O(\kappa^2)$ is diminished by the limit $\kappa\rightarrow0^+$ above, leaving the dominant $\kappa$-linear term.
This property is model-independent although, in the example before, we have used a model-dependent scaling invariance~(\ref{scaling}) to manifest it.

However,
the converse is generally not true.
Namely,
given a multiplicative reward,
its optimal policy cannot be approached with an arbitrary precision
by any other additive reward.
Our exactly solvable case in this work shows that the multiplicative reward, in the viewpoint of the additive one, has a long-range correlation, i.e., the reward function 
$\mathcal{R}_\cdot^t$ in Eq.~(\ref{reward_m}) by a Taylor expansion contains terms like $s_ts_{t+k}$ for arbitrarily large $k<T-t$. 
This non-perturbative nature cannot be captured by any additive scheme.
It exactly means that the multiplicative reward is a more general framework than the additive reward due to an additional free parameter $\kappa$ to adjust the weight between reward and other model factors, e.g., the noise.

In a short summary,
if the real world indeed prefers the additive reward to produce a better policy,
we can still use the multiplicative reward by tuning $\kappa$ to a smaller value during the series of experiments and tests.
On the other side,
if the multiplicative way is preferable in the real system, 
the additive accumulation of reward generically cannot give a satisfying optimal policy.

\section{Conclusion}
\label{sec_con}
In this work, we propose a new multiplication way of reward accumulations and develop a rigorous solution in a linear transition model. 
In contrast to the conventional additive reward case, 
our optimal policy is explicitly dependent on the noise of linear transition models.
Furthermore, 
we also show that the multiplicative scheme is a general framework that covers the additive one with an arbitrary precision.
We expect that our proposal of extension of the reward accumulation can have a wide application in real systems. 

\acknowledgments
The authors are grateful to Professor Koji Tsuda for helpful advice on the manuscript.
Y.~Y. was supported by JSPS fellowship and X.~S. was supported by the China Scholarship Council.
This work was supported in part by MEXT/JSPS KAKENHI Grant No. JP19J13783 (Y.~Y.) and CSC No. 201809120018 (X.~S.).

%

\end{document}